\documentclass[11pt]{article}

\usepackage[final]{acl}

\usepackage{times}
\usepackage{latexsym}
\usepackage[T1]{fontenc}
\usepackage[utf8]{inputenc}
\usepackage{microtype}
\usepackage{inconsolata}
\usepackage{graphicx}

\usepackage{amsmath}
\usepackage{amssymb}
\usepackage{algorithm}
\usepackage{algpseudocode}
\usepackage{cleveref}
\usepackage{bbm}
\usepackage{comment}
\usepackage{booktabs}
\usepackage{subcaption}

\usepackage{multirow}
\usepackage{colortbl}
\title{MDP-GRPO: Stabilized Group Relative Policy Optimization for Multi-Constraint Instruction Following}

\author{
	\textbf{Mohammad~Mahdi Salmani-Zarchi\textsuperscript{1}\thanks{\ \ Corresponding authors.}},
	\textbf{Zahra~Rahimi\textsuperscript{2}},
	\textbf{Heshaam~Faili\textsuperscript{1}},
	\textbf{Mohammad~Javad~Dousti\textsuperscript{1,*}} \\
	\\
	\textsuperscript{1}Department of Electrical and Computer Engineering, College of Engineering, University of Tehran, Tehran, Iran \\
	\textsuperscript{2}Department of Statistics, Mathematics and Computer Science, Allameh Tabataba'i University, Tehran, Iran \\
	\\
	\texttt{m.salmani78@ut.ac.ir} \quad 
    \texttt{za.rah@atu.ac.ir} \quad
    \texttt{hfaili@ut.ac.ir} \quad
    \texttt{mjdousti@ut.ac.ir}
}

\begin{document}
\maketitle
\begin{abstract}
Reinforcement learning with verifiable rewards is ideal for multi-constraint instruction following, yet standard group-relative policy optimization (GRPO) becomes unstable under discrete, low-dispersion rewards, where within-group reward distributions are frequently homogeneous.
We identify and formalize three pathologies of z-score group normalization in this regime: low-variance amplification, mean-centering blindness, and zero-variance collapse.
To address them, we propose MDP-GRPO, which stabilizes learning through (1) multi-temperature sampling to increase reward dispersion, (2) dual-anchor advantages to restore gradients in homogeneous groups and stop mean-centering blindness, (3) prospect-theoretic shaping to bound updates and penalize violations based on Kahneman \& Tversky's theory, and (4) asymmetric KL regularization.
Evaluated on FollowBench, IFEval, and a curated multi-constraint dataset, MDP-GRPO outperforms standard GRPO, improving strict constraint satisfaction by up to 5.0\% on Llama-3.2-3B.
Our method also enables stable convergence with small group sizes while preserving general capabilities on MMLU and ARC
\footnote{Our codes are available at \url{https://github.com/m-salmani78/MDP-GRPO}}.
\end{abstract}

\begin{figure}[t]
\includegraphics[trim={0cm 0.2cm 0cm 0cm}, width=\linewidth]{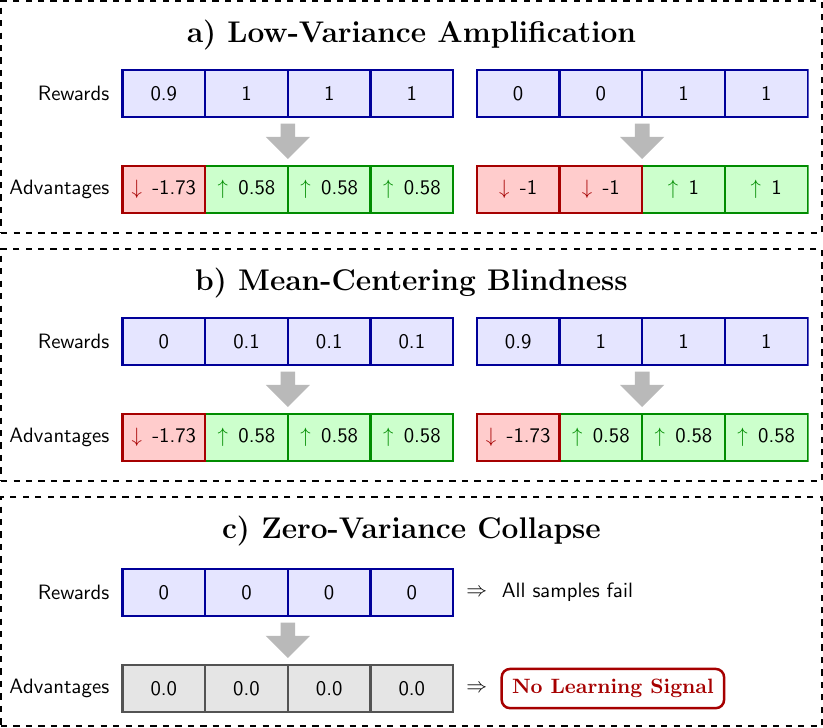}
\centering
\caption{Illustration of group-normalized advantage pathologies in GRPO.}
\label{fig:grpo_pathologies}
\end{figure}

\section{Introduction}

Large language models (LLMs) can follow many natural-language instructions \citep{ouyang2022instructgpt,chung2022instructiontuning}.
Yet, they remain brittle when a request bundles multiple explicit constraints, such as asking the LLM to respond in a particular structure with an exact ending phrase, while adhering to strict lexical constraints and casing rules \citep{jiang2024followbench,geng2023gcd,park2025gcd}.
In real deployments, these \emph{multi-constraint} prompts are common: product and legal templates demand fixed formats \citep{westermann2024dallma,narendra2024contracts}, developer tools require machine-readable outputs \citep{schick2023toolformer,yao2023react,shen2025slot}, and safety constraints impose hard exclusions \citep{chen2025safetyconstraints}. In such settings, a response that is mostly correct but violates a single constraint is often unusable \citep{zhou2023ifeval}.

Recently, reinforcement learning with \emph{verifiable} rewards (RLVR) has emerged as a promising direction, where each constraint is checked deterministically and the model is optimized to satisfy as many constraints as possible \citep{wen2025rlvr,guo2025deepseekr1}.
Relying on rule-based checkers is a deliberate design choice for domains requiring strict compliance; it provides a highly reliable learning signal, avoids expensive preference labels, and completely eliminates the bias and hallucinations inherent in learned reward models or LLM-as-a-judge evaluators \citep{chen2024judgebias,ye2025justiceprejudice}.
However, the resulting rewards are discrete and frequently low-variance early in training, making stable policy-gradient learning notoriously challenging \citep{wen2025rlvr,shapingsparse2025}.

In particular, when policy updates rely on group-relative normalization (as in GRPO-style methods) \citep{shao2024deepseekmath}, multi-constraint reward structures induce three recurring pathologies (\Cref{fig:grpo_pathologies}).
First, \textbf{low-variance amplification}: when within-group reward variance is small but nonzero, z-score normalization can inflate minor reward differences into disproportionately large advantages, yielding brittle updates.
Second, \textbf{mean-centering blindness}: because z-score normalization discards absolute reward level, semantically distinct groups (e.g., consistently easy vs.\ consistently hard prompts) can receive nearly identical normalized advantage patterns, obscuring which cases truly require correction. %\citep{mapo2025}.
Third, \textbf{zero-variance collapse}: homogeneous groups arise frequently early in training, and if all samples for a prompt satisfy or violate the same constraints, rewards will be equal and thereby group-normalized advantages collapse to zero, providing no learning signal.
The resulting advantages can destabilize training and even trigger regressions in general capabilities \citep{lin2023alignmenttax,kotha2024forgetting}.

This paper introduces three independent and composable modules to make group-based RL reliable in these failure modes.
We target both prevention and treatment.
To prevent homogeneous groups, we employ \emph{multi-temperature sampling}, mixing high-quality and exploratory completions to ensure within-group reward dispersion.
To treat signal collapse and absolute reward blindness, we introduce \emph{dual-anchor advantages} which interpolate between group-relative and goal-aware advantages.
We then apply a bounded, asymmetric shaping function inspired by \emph{Prospect Theory} \citep{kahneman1979prospect} to yield more human-aligned update magnitudes.
While our primary focus is on deterministic and verifiable constraints, the core components of our approach are fundamentally agnostic to the reward source. They can be readily adapted to reward models or human-feedback settings.

We evaluate on our multi-constraint test set as well as established benchmarks (\textsc{IFEval} and \textsc{FollowBench}). We additionally track \textsc{MMLU} and \textsc{ARC} to verify that instruction-following gains do not come at the expense of degrading general knowledge and reasoning capabilities \citep{hendrycks2021mmlu,clark2018arc}.

Overall, our contributions are:
(1) We identify three failure modes of group-relative policy updates: low-variance amplification, mean-centering blindness, and zero-variance collapse; and we show how these arise naturally from discrete, low-dispersion reward structures.
(2) We propose multi-temperature group sampling as a prevention mechanism that improves reward diversity in small-batch sampling (e.g., $G=4$) without altering the underlying objective.
(3) We introduce dual-anchor advantages to restore the learning signal in zero-variance and mean-centering blindness groups.
(4) We apply bounded, asymmetric advantage shaping inspired by Prospect Theory to cap update magnitudes and emphasize violations, thereby improving robustness.
(5) We evaluate the method on a custom multi-constraint test set, \textsc{IFEval}, and \textsc{FollowBench}, demonstrating that our approach improves constraint satisfaction and stability with negligible overhead, while retaining general capabilities.

\begin{figure*}[t]
\centering
\includegraphics[trim={0.2cm 0.4cm 0.2cm 0.6cm}, width=\textwidth]{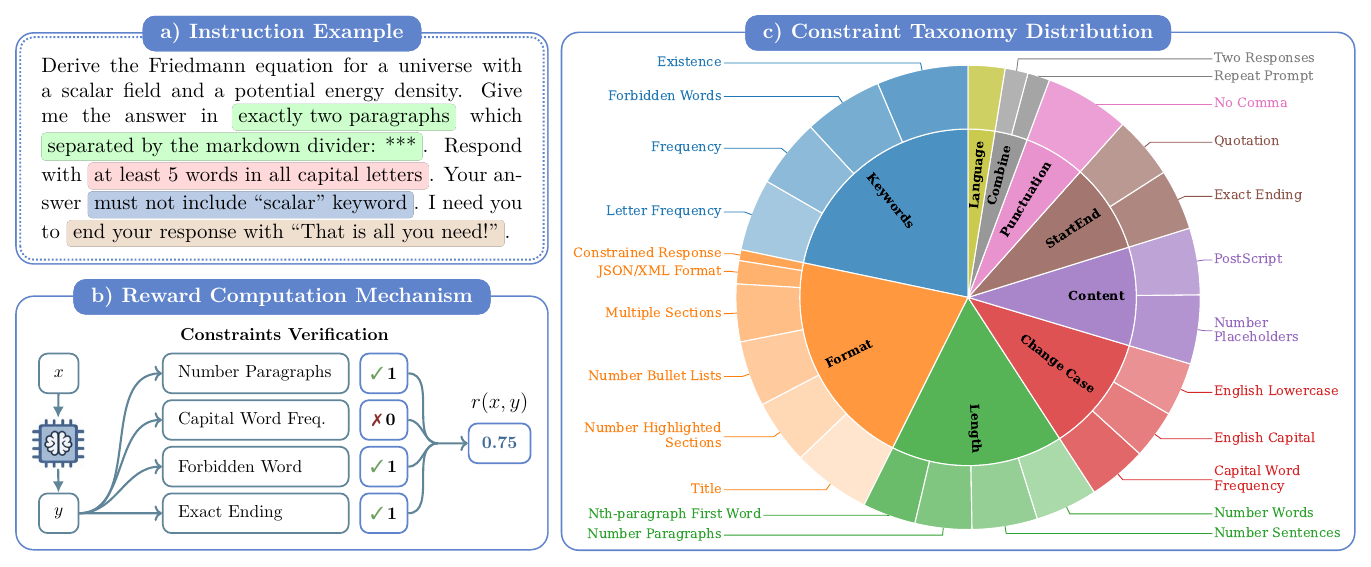}
\caption{Characteristics of the dataset. (a) Example of a multi-constraint instruction. (b) Reward computation mechanism via rule-based constraint verification. (c) Taxonomy and relative frequency of constraint types.}
\label{fig_data_overview}
\end{figure*}
% ==========================================================
\section{Related Work}
\label{sec:related}

\subsection{Multi-constraint instruction data}
Multi-constraint instruction following has been studied via synthetic constraint injection and large-scale data construction.
\citet{he-etal-2024-complex} show that training on compositional constraints improves complex instruction adherence and transfers to novel constraint combinations.
RECAST further expands coverage with large, automatically verifiable multi-constraint data and validators that support both supervised training and RL with verifiable signals \citep{liu2025recast}.
We evaluate on established instruction-following benchmarks alongside a constraint-heavy test set \citep{zhou2023ifeval,jiang2024followbench}.

\subsection{RLVR and group-relative optimization}
Reinforcement learning with verifiable rewards replaces preference modeling with deterministic checks and has recently enabled strong gains on reasoning-style tasks \citep{wen2025rlvr,guo2025deepseekr1}.
Group-based optimizers such as GRPO estimate advantages from multiple samples per prompt via within-group normalization, avoiding an explicit critic \citep{shao2024deepseekmath}.
In discrete and partially sparse reward regimes, however, group normalization can become fragile under low-dispersion or homogeneous groups, motivating modifications to sampling and advantage estimation.

\subsection{Robust group advantages}
Recent work addresses pathologies of group-relative advantages.
MAPO proposes mixed, sample-adaptive advantage schemes to mitigate advantage allocation failures \citep{mapo2025}.
NGRPO targets all-negative (homogeneously incorrect) groups and recovers learning signal via advantage calibration and asymmetric clipping \citep{nan2025ngrpo}.
Our approach is complementary: we reduce homogeneous groups through temperature-diverse sampling, restore goal-aware signal with dual-anchor advantages, and apply bounded, loss-averse shaping to stabilize updates under multi-constraint RLVR.

\subsection{Prospect-style, risk-aware objectives}
Prospect-theoretic objectives have been explored for alignment, notably in Kahneman-Tversky Optimization (KTO) \citep{ethayarajh2024kto}, which incorporates loss aversion via a human-centric utility function \citep{kahneman1979prospect}.
KTO applies this framework at the \emph{objective level} to offline, un-paired preference data (e.g., binary thumbs-up/down labels). 
In contrast, our approach applies prospect-inspired shaping strictly at the \emph{advantage} level for group-based RL, yielding bounded, asymmetric updates while keeping the underlying reward specification unchanged.

% ==========================================================
\section{Problem Setup}
\label{sec:setup}

\subsection{Task Definition}
\label{subsec:task_definition}

Given an instruction $x$, a model generates a completion $y$.
Each instruction specifies $C(x)$ explicit constraints, each of which is validated by a deterministic rule-based checker.

The score of a completion, $r(x,y) \in [0, 1]$, is defined by the fraction of satisfied constraints.
Let $c_t(x,y) \in \{0, 1\}$ be the binary verification result of the $t$-th constraint. The resulting reward is:
\begin{equation}
    r(x,y) = \frac{1}{C(x)}\sum_{t=1}^{C(x)} c_t(x,y).
\end{equation}

This reward is used for RL training. We focus on \emph{critic-free} group-based objectives, specifically Group Relative Policy Optimization (GRPO), which avoids the computational overhead of training a separate value network.

\subsection{Multi-Constraint Instruction-Following Dataset}
\label{subsec:dataset}

To study instruction following under explicit compositional constraints, we compile a dataset of 3,000 training prompts.

\subsubsection{Seed Prompts}
We begin with a collection of seed instructions sourced through a hybrid approach.
Approximately one-third of the seed prompts are sampled from the dataset introduced by \citet{he-etal-2024-complex}.
The remaining prompts are newly curated to ensure a balanced coverage across three broad intent types: general Q\&A, creative writing, and material assistance.
All seed prompts are in English and represent the base request without explicit constraint markup, resembling natural user queries.

\subsubsection{Constraint Taxonomy}

We adopt the constraint taxonomy from prior work on complex instruction synthesis~\citep{he-etal-2024-complex}.
We utilize 26 constraint types organized into 9 high-level categories (see \Cref{fig_data_overview}).

Each constraint type is instantiated through parameterized templates (e.g., varying forbidden words or numerical limits). Crucially, all constraints are designed to be automatically verifiable using deterministic validators (regular expressions, parsers), enabling fast, reproducible reward computation.

\subsubsection{Constraint Injection Procedure}
Given a seed instruction $x$, we inject between 1 and 6 constraints. To ensure validity, we enforce strict compatibility rules (e.g., preventing contradictory casing constraints).
We filter constraints that conflict with those already selected, then render the final instruction by appending natural-language descriptions of the chosen constraints to the seed prompt. The underlying formal constraint set $\{c_t\}_{t=1}^{C(x)}$ is stored for evaluation. Constraint counts follow a unimodal distribution peaking at 3--5 constraints, emphasizing moderately complex compositions.

\subsection{Baseline: Group Relative Policy Optimization}
\label{subsec:grpo_baseline}

We adopt GRPO \citep{shao2024deepseekmath} as our baseline.
For each query $x$, GRPO samples a group of $G$ outputs $\{y_1, y_2, \dots, y_G\}$ from the old policy $\pi_{\theta_{old}}$. The policy is optimized to maximize the surrogate objective:
\begin{align}
\mathcal{J}(\theta) &= \mathbb{E}_{x \sim \mathcal{D}, \{y_i\} \sim \pi_{\theta_{old}}} \left[
\right. \notag \\ 
&\quad \left.
\frac{1}{G} \sum_{i=1}^G \min 
\left( \frac{\pi_\theta(y_i|x)}{\pi_{\theta_{old}}(y_i|x)} A_i, \text{clip}(\dots) A_i \right)
\right. \notag \\ 
&\quad \left.
- \beta_{KL} \mathbb{D}_{KL} \right],
\end{align}
where $\beta_{KL}$ controls the KL-divergence penalty.
Crucially, GRPO computes the advantage $A_i$ for the $i$-th completion using group-relative z-score normalization:
\begin{equation}
    \label{eq:standard_grpo_adv}
    A_i = \frac{r_i - \mu_{\text{group}}}{\sigma_{\text{group}} + \epsilon},
\end{equation}
where $r_i = r(x, y_i)$ is the reward of completion $y_i$, while $\mu_{\text{group}}$ and $\sigma_{\text{group}}$ are the mean and standard deviation of rewards within the sampled group.
%It is this exact normalization mechanism (Eq.~\ref{eq:standard_grpo_adv}) that leads to instability when applied to the discrete, low-variance rewards characteristic of multi-constraint tasks.

\begin{figure*}[t]
\centering
\includegraphics[trim={0.5cm 0.5cm 0.5cm 0.5cm}, width=\textwidth]{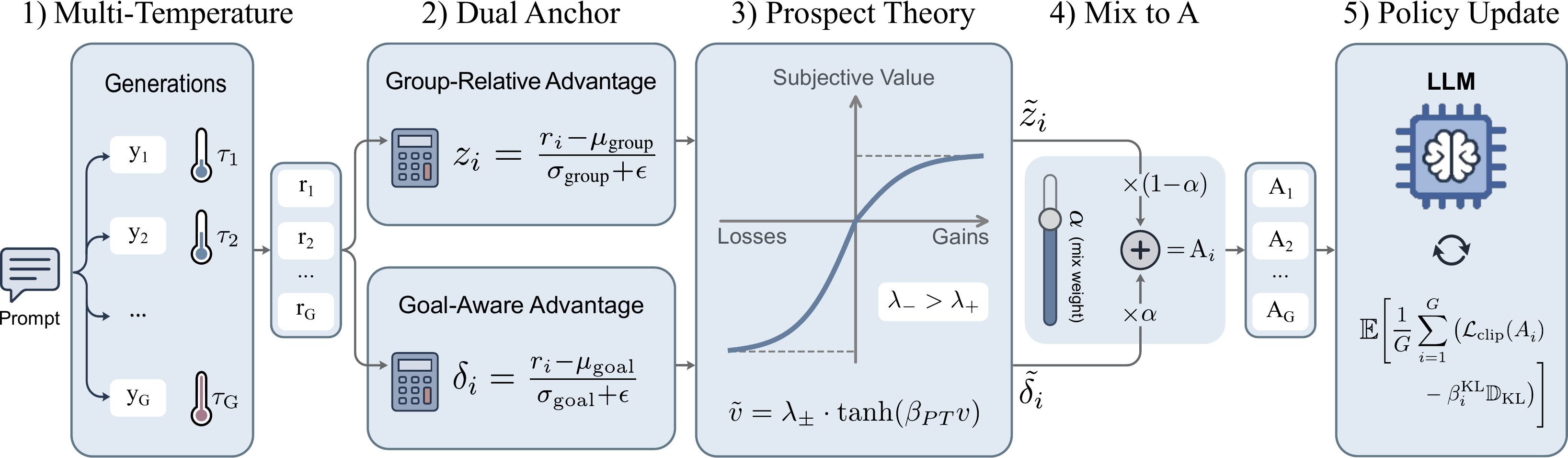}
\caption{MDP-GRPO pipeline. (1) Multi-temperature sampling generates a group of $G$ completions per prompt.
(2) Compute group-relative $z_i$ and goal-aware $\delta_i$ advantage signals.
(3) Prospect-theoretic shaping yields $\tilde{z_i}$ and $\tilde{\delta_i}$.
(4) Mixing $\tilde{z_i}$ and $\tilde{\delta_i}$ gives $A_i$.
(5) Apply a standard GRPO policy update using $A_i$.}
\label{fig_pipeline_overview}
\end{figure*}
% ==========================================================
\section{Method}
\label{sec:method}

We introduce MDP-GRPO (visualized in \Cref{fig_pipeline_overview}), a method designed to stabilize policy gradients for multi-constraint tasks.
Our approach addresses the pathologies of standard group-relative advantages through three complementary mechanisms:
(1) Temperature-diverse sampling to prevent zero-variance groups;
(2) Dual-anchor advantages to maintain signal when group variance collapses; and
(3) Prospect-theoretic shaping to bound updates and enforce loss aversion.

\subsection{Motivation: Pathologies of Group Normalization}
\label{subsec:method_motivation}
As defined in \Cref{subsec:grpo_baseline}, standard GRPO relies on the z-scored advantage $z_i = (r_i - \mu_{\text{group}}) / (\sigma_{\text{group}} + \epsilon)$.
In the multi-constraint regime, this formulation suffers from three specific failure modes:
\begin{enumerate}
    \item \textbf{Low-Variance Amplification:} When $\sigma_{\text{group}}$ is small but non-zero, minor reward fluctuations are inflated into disproportionately large advantages, causing brittle updates.
    \item \textbf{Zero-Variance Collapse:} When all completions in a group satisfy the exact same subset of constraints (a common occurrence with deterministic checkers), $\sigma_{\text{group}} \to 0$. The advantage becomes undefined or noise-dominated (determined solely by $\epsilon$), providing no valid learning signal.
    \item \textbf{Mean-Centering Blindness:} A group where the model fails no constraints and a group where it fails all constraints can produce identical normalized advantages, discarding critical absolute performance information.
\end{enumerate}

\subsection{Temperature-Diverse Group Sampling}
\label{sec:temp}

To mitigate the frequency of homogeneous groups, we modify the sampling strategy. Instead of sampling all $G$ outputs with a fixed temperature, we utilize a temperature schedule $\boldsymbol{T}=[\tau_1, \dots, \tau_G]$.

For the $i$-th element of the group, we sample $y_i \sim \pi_{\theta_{\text{old}}}(\cdot\mid x;\tau_i)$.
We configure $\boldsymbol{T}$ to purposefully mix exploitation and exploration, e.g., $\boldsymbol{T} = [0.1, 0.4, 0.7, 1.0]$. Lower temperatures encourage exploitation by stabilizing a high-quality, ``best-effort'' baseline, while higher temperatures induce exploration, increasing the likelihood of within-group reward dispersion without altering the underlying group-based RL objective.

\subsection{Dual-Anchor Advantages}
\label{sec:dualanchor}

To address mean-centering blindness and zero-variance collapse, we introduce a goal-aware anchor.
We model a \emph{neutral} baseline as a binomial process where each of the $C(x)$ constraints is satisfied independently with probability $p=0.5$.
For a reward normalized to $[0,1]$, this neutral baseline has mean $\mu_{\text{goal}} = 0.5$ and standard deviation $\sigma_{\text{goal}} = \frac{1}{2\sqrt{C(x)}}$.
We define the absolute, goal-aware advantage $\delta_i$ as:
\begin{equation}
    \delta_i = 2\sqrt{C(x)}(r_i - 0.5).
    \label{eq:delta}
\end{equation}
This scales the reward such that a completion satisfying all constraints yields $\delta_i = +\sqrt{C(x)}$, and one satisfying none yields $\delta_i = -\sqrt{C(x)}$. Crucially, $\delta_i$ remains well-defined and informative even when the group variance is zero, providing our second anchor alongside the standard group-relative advantage $z_i$.

\subsection{Prospect-Theoretic Advantage Shaping}
\label{sec:prospect}

Prospect Theory \citep{kahneman1979prospect} is a foundational model in behavioral economics that describes how humans make decisions involving risk. It demonstrates a distinct, non-linear relationship between objective outcomes and their perceived subjective value.
A core tenet of this theory is \emph{loss aversion}: the psychological pain of losing 50 dollars is experienced much more intensely than the satisfaction of winning 50 dollars. Additionally, human perception exhibits diminishing sensitivity; as the absolute magnitude of gains or losses increases, marginal changes are felt less strongly.

In the context of GRPO, the advantage signals act as these objective outcomes. Standard policy gradient updates treat advantages linearly.
However, raw standard advantages can grow unboundedly (especially when the batch variance $\sigma$ is extremely small).
Furthermore, treating positive and negative completions symmetrically fails to adequately penalize the model when it regresses on strict constraints.

Inspired by Prospect Theory, we frame the raw advantage as an objective outcome that must be transformed into a subjective, "human-aligned" value signal. We apply a bounded, asymmetric shaping function to our raw advantage signals before mixing them.

Let $v_i \in \{z_i, \delta_i\}$ represent either of the raw advantage signals. We approximate the prospect-theoretic value function using a scaled `tanh' transformation to compute the shaped advantage $\tilde{v}_i$:
\begin{equation}
    \tilde{v}_i =
    \begin{cases}
    \lambda_{+}\tanh(\beta_{\text{PT}} v_i) & \text{if } v_i \ge 0 \\
    \lambda_{-}\tanh(\beta_{\text{PT}} v_i) & \text{if } v_i < 0
    \end{cases}
    \label{eq:prospect}
\end{equation}
By setting $\lambda_{-}>\lambda_{+}>0$, we enforce loss aversion: negative outcomes (constraint violations) result in larger magnitude gradient updates than equivalent positive outcomes, heavily penalizing regression. Concurrently, the $\tanh$ function captures diminishing sensitivity and strictly bounds the advantages to $[-\lambda_-, \lambda_+]$, preventing unbounded gradients and stabilizing training. Applying this function yields the shaped group-relative advantage $\tilde{z}_i$ and the shaped goal-aware advantage $\tilde{\delta}_i$.

Finally, we compute the final combined advantage $A_i$ by interpolating between the two shaped signals:
\begin{equation}
    A_i = (1-\alpha) \tilde{z}_i + \alpha \tilde{\delta}_i,
    \label{eq:mix_advantages}
\end{equation}
where $\alpha \in [0,1]$ controls the reliance on the absolute anchor.

\paragraph{Theoretical Validity.}
Crucially, because the mixed baseline relies only on the group mean and a fixed goal-aware anchor, it remains action-independent for any individual sample within the group. Therefore, its incorporation is theoretically valid under the policy gradient theorem \citep{sutton1999policy}. While this introduces a controlled bias relative to the standard zero-mean GRPO estimator, this goal-aware bias is essential for stabilizing gradients in low-dispersion regimes where group normalization becomes ill-conditioned (i.e., as $\sigma_{\text{group}} \to 0$). We discuss and empirically validate this bias-variance tradeoff, alongside the sensitivity of the mixing weight $\alpha$, in Appendix~\ref{sec:appendix_da_ablations}.

\subsection{MDP-GRPO Objective and Algorithm}
\label{sec:pg}

We integrate these components into the GRPO framework. We also optionally employ an \emph{asymmetric KL penalty} to permit larger deviations when the model is improving (positive advantage) while strictly constraining it when performance drops.

The final objective is:
\begin{align}
\mathcal{J}(\theta) &= \mathbb{E} \left[ \frac{1}{G} \sum_{i=1}^G \left( \mathcal{L}_{\text{clip}}(A_i) 
- \beta_i^{\text{KL}} \mathbb{D}_{\text{KL}} \right) \right],
\label{eq:objective}
\end{align}
where $\mathcal{L}_{\text{clip}}$ is the standard GRPO clipping term using $A_i$, and the asymmetric KL coefficient is:
\begin{equation}
    \beta_i^{\text{KL}} = \begin{cases} \beta_{\text{low}} & \text{if } A_i \ge 0 \\ \beta_{\text{high}} & \text{if } A_i < 0 \end{cases}
\end{equation}

The full training procedure is detailed in Algorithm \ref{alg:pg} in \Cref{sec:appendix_alg}.
% TODO: Say these thre components are independent and could be disabled.

% ==========================================================
\section{Experiments}
\label{sec:experiments}

\begin{table*}[t]
\centering
\small
\setlength{\tabcolsep}{5.5pt}
\begin{tabular}{lcc|cc|cc||cc|cc|cc}
\toprule
\multirow{1}{*}{\textbf{Models (G=8)}} & \multicolumn{6}{c}{\textbf{Gemma-2-2B-Instruct}} & \multicolumn{6}{c}{\textbf{Llama-3.2-3B-Instruct}} \\
\cmidrule(lr){2-7}\cmidrule(lr){8-13}
\multirow{2}{*}{\textbf{Benchmarks}} & \multicolumn{2}{c}{\textbf{IFEVAL}} & \multicolumn{2}{c}{\textbf{Custom Test Set}} & \multicolumn{2}{c}{\textbf{FollowBench}} & \multicolumn{2}{c}{\textbf{IFEVAL}} & \multicolumn{2}{c}{\textbf{Custom Test Set}} & \multicolumn{2}{c}{\textbf{FollowBench}} \\
\cmidrule(lr){2-3}\cmidrule(lr){4-5}\cmidrule(lr){6-7}\cmidrule(lr){8-9}\cmidrule(lr){10-11}\cmidrule(lr){12-13}
 & \textbf{SSR} & \textbf{HSR} & \textbf{SSR} & \textbf{HSR} & \textbf{SSR} & \textbf{HSR} & \textbf{SSR} & \textbf{HSR} & \textbf{SSR} & \textbf{HSR} & \textbf{SSR} & \textbf{HSR} \\
\midrule
Baseline 
& 56.7 & 45.1 & 54.8 & 18.8 & 63.7 & 52.9 
& 54.2 & 46.8 & 60.3 & 20.8 & 69.7 & 59.8 \\
GRPO 
& 73.7 & 62.4 & 68.4 & 29.0 & 64.0 & 53.2 
& 66.1 & 58.5 & 65.1 & 24.8 & 68.4 & 58.9 \\

\midrule
MT-GRPO 
& 73.9 & 63.7 & 68.6 & 29.6 & 64.3 & 53.7 
& 68.1 & 59.1 & 65.4 & 24.6 & 70.0 & 59.9 \\
DA-GRPO 
& 75.2 & 64.7 & 70.7 & 32.6 & 65.4 & 56.1 
& 71.2 & \textbf{62.5} &  66.1 & 24.9 & 69.6 & 58.9 \\
PT-GRPO 
& \textbf{75.7} & \textbf{65.8} & 71.4 & 30.6 & 66.5 & 56.1 
& 70.1 & 61.3 & 64.7 & 23.6 & 69.3 & 57.6 \\

\midrule
DA-PT-GRPO 
& 74.8 & 64.3 & \textbf{70.8} & \textbf{33.4} & \textbf{68.2} & \textbf{59.7} 
& \textbf{71.5} & 60.4 & \textbf{66.3} & \textbf{25.4} & \textbf{70.2} & 59.2 \\
\midrule
\rowcolor{gray!20}
MDP-GRPO
& 75.3 & 64.1 & 70.3 & 32.8 & 66.9 & 57.4 
& 71.3 & 59.8 & 65.8 & 25.2 & 69.4 & 59.1 \\
\bottomrule
\end{tabular}
\caption{Instruction-following performance of GRPO variants using a standard group size ($G=8$). Results are reported using SSR and HSR across three benchmarks.}
\label{tab:if_eval_ssr_hsr}
\end{table*}

\begin{table*}[t]
\centering
\small
\setlength{\tabcolsep}{5.5pt}
\begin{tabular}{lcc|cc|cc||cc|cc|cc}
\toprule
\multirow{1}{*}{\textbf{Models (G=4)}} & \multicolumn{6}{c}{\textbf{Gemma-2-2B-Instruct}} & \multicolumn{6}{c}{\textbf{Llama-3.2-3B-Instruct}} \\
\cmidrule(lr){2-7}\cmidrule(lr){8-13}
\multirow{2}{*}{\textbf{Benchmarks}} & \multicolumn{2}{c}{\textbf{IFEVAL}} & \multicolumn{2}{c}{\textbf{Custom Test Set}} & \multicolumn{2}{c}{\textbf{FollowBench}} & \multicolumn{2}{c}{\textbf{IFEVAL}} & \multicolumn{2}{c}{\textbf{Custom Test Set}} & \multicolumn{2}{c}{\textbf{FollowBench}} \\
\cmidrule(lr){2-3}\cmidrule(lr){4-5}\cmidrule(lr){6-7}\cmidrule(lr){8-9}\cmidrule(lr){10-11}\cmidrule(lr){12-13}
 & \textbf{SSR} & \textbf{HSR} & \textbf{SSR} & \textbf{HSR} & \textbf{SSR} & \textbf{HSR} & \textbf{SSR} & \textbf{HSR} & \textbf{SSR} & \textbf{HSR} & \textbf{SSR} & \textbf{HSR} \\
\midrule
Baseline & 56.7 & 45.1 & 54.8 & 18.8 & 63.7 & 52.9 & 54.2 & 46.8 & 60.3 & 20.8 & 69.7 & 59.8 \\
GRPO & 69.7 & 58.2 & 67.3 & 28.6 & 64.5 & 53.5 & 67.2 & 55.0 & 63.3 & 21.6 & 69.4 & 60.5 \\
\midrule
MT-GRPO & 71.1 & 59.4 & 68.4 & \textbf{30.6} & 64.2 & 53.5 & \textbf{70.5} & \textbf{58.4} & 63.4 & \textbf{22.8} & 69.9 & \textbf{60.7} \\
DA-GRPO & 70.5 & 58.9 & 67.3 & 28.5 & 64.6 & 53.6 & 69.3 & 56.5 & 63.5 & 22.6 & 68.8 & 59.6 \\
PT-GRPO & 69.2 & 56.6 & \textbf{68.7} & 29.9 & \textbf{66.4} & \textbf{55.7} & 69.7 & 58.2 & 63.1 & 22.0 & \textbf{71.1} & 61.6 \\
\midrule
DA-PT-GRPO & 70.1 & 58.9 & 67.2 & 29.4 & 63.9 & 53.4 & 68.5 & 56.9 & \textbf{64.0} & 22.6 & 69.4 & 59.2 \\
\midrule
\rowcolor{gray!20}
MDP-GRPO & \textbf{71.2} & \textbf{59.5} & 67.8 & 30.4 & 65.8 & 55.3 & 68.5 & 57.0 & 63.7 & \textbf{22.8} & 70.2 & 60.7 \\
\bottomrule
\end{tabular}
\caption{Instruction-following performance of GRPO variants using a reduced group size ($G=4$). Results are reported using SSR and HSR across three benchmarks.}
\label{tab:g4_results}
\end{table*}

\subsection{Benchmarks}
\label{sec:benchmarks}
We evaluate instruction-following on two standard benchmarks: \textsc{IFEval} \citep{zhou2023ifeval}, which uses strict code-based verification, and \textsc{FollowBench} \citep{jiang2024followbench}, which utilizes a hybrid evaluation framework (rules and LLM judges) to assess fine-grained constraints.
To complement these public benchmarks, we additionally report results on our custom test set of 500 multi-constraint prompts, created using a workflow similar to the one described in \Cref{subsec:dataset}.

\subsection{Evaluation Metrics}
\label{sec:metrics}

Following established protocols \citep{zhou2023ifeval,jiang2024followbench}, we report two complementary accuracy metrics.
Let $N$ be the number of evaluation prompts. For the $i$-th prompt with $C_i$ constraints, let $c_{i,j} \in \{0,1\}$ denote the satisfaction status of the $j$-th constraint.

\paragraph{Constraint Accuracy (SSR).}
We define \textit{Soft Success Rate} (SSR) as the mean constraint satisfaction rate across all instructions, analogous to $\text{acc}_{\text{con}}$ in prior work. Formally, $\text{SSR} = \frac{1}{N} \sum_{i=1}^{N} \left( \frac{1}{C_i}\sum_{j=1}^{C_i} c_{i,j} \right)$. This metric captures partial progress under sparse multi-constraint rewards.

\paragraph{Prompt Accuracy (HSR).}
We define \textit{Hard Success Rate} (HSR) as the strict all-constraints success rate, analogous to $\text{acc}_{\text{ins}}$. It requires satisfying every constraint within a prompt simultaneously: $\text{HSR} = \frac{1}{N} \sum_{i=1}^{N} \prod_{j=1}^{C_i} c_{i,j}$. This reflects whether an output is fully usable in strict template-like settings.

\subsection{Models}
\label{sec:models}
We conduct experiments on two instruction-tuned models: Gemma-2-2B-Instruct and Llama-3.2-3B-Instruct.
Unless otherwise stated, the policy is initialized from the corresponding instruct model, and the reference policy $\pi_{\text{ref}}$ used for KL regularization is this frozen initialization checkpoint.

\subsection{Compared Methods}
\label{sec:methods_compared}
To isolate the contribution of each module, we compare six training variants. We evaluate the zero-shot Baseline (the initial instruct model) and standard GRPO against our individual ablations:
MT-GRPO (Multi-Temperature sampling), DA-GRPO (Dual-Anchor advantages), PT-GRPO (Prospect-Theoretic shaping), and DA-PT-GRPO (combining DA and PT).
Finally, we evaluate MDP-GRPO, our full pipeline combining all three mechanisms to address stability and performance in multi-constraint regimes.

\subsection{Training Setup}
\label{sec:training_setup}
We primarily train with a group size of $G=8$ completions per prompt. We also perform ablations with $G=4$ to demonstrate the efficacy of multi-temperature sampling in small-batch regimes.
We train all models on a single NVIDIA A100 GPU.

\paragraph{Decoding.}
We use stochastic decoding with \texttt{do\_sample=True}. For MT variants, the group is sampled using a fixed temperature schedule $\boldsymbol{T}=[\tau_1,\dots,\tau_G]$. For standard methods, we use a fixed temperature.
We use \texttt{top\_p} nucleus sampling with $p=0.9$ and a maximum generation length of $1024$ tokens.

\paragraph{Optimization.}
We use a learning rate of $1\times 10^{-5}$, batch size $32$, PPO clip $\epsilon_{\text{clip}}=0.2$, and a base KL coefficient $\beta_{\text{KL}}=0.01$.
For runs utilizing asymmetric KL, we set $\beta_{\text{KL}}^{\text{high}}=0.025$ (applied when advantages are negative) and $\beta_{\text{KL}}^{\text{low}}=0.01$.
Models are trained for one epoch.

\paragraph{Method Hyperparameters.}
For dual-anchor variants, we evaluate static mixing weights $\alpha \in \{0.1, 0.2, 0.4\}$.
Unless otherwise noted, we use $\alpha=0.2$ and set the anchor mean $\mu_{\text{goal}} = 0.5$ (neutral baseline).
For Prospect Shaping variants, we set $\beta_{\text{PT}}=0.8$ to approximate empirical findings in behavioral economics, and $(\lambda_{+}, \lambda_{-}) = (1.25, 2.0)$. This configuration ensures a slope of $\approx 1.0$ for positive advantages.

\paragraph{Ablations and Sensitivity.}
We report additional sensitivity analyses for the dual-anchor design (mixing weight $\alpha$ and the goal-aware center in $\delta_i$) in \Cref{sec:appendix_da_ablations}.
Ablations for prospect-theoretic shaping and asymmetric KL regularization are reported in \Cref{sec:appendix_pt_ablations}.

\subsection{General Capability Evaluation}
\label{sec:general_capabilities}
To assess alignment tax, we evaluate all trained policies on \textsc{MMLU} \citep{hendrycks2021mmlu} and \textsc{ARC} (Easy and Challenge) \citep{clark2018arc} in \Cref{sec:results_general}.
% TODO: Update the table of MMLU and ARC

% ==========================================================
\section{Results}
\label{sec:results}

\subsection{Main Results on Instruction Following}
\label{subsec:main_results}

Table~\ref{tab:if_eval_ssr_hsr} reports instruction-following performance under our standard group size ($G=8$).
Across Gemma-2-2B-Instruct and Llama-3.2-3B-Instruct, MDP-GRPO variants generally improve over both the baseline model and standard GRPO on IFEval, FollowBench, and our Custom Test Set, with the largest gains appearing on harder multi-constraint prompts.

On IFEval, the individual components exhibit mechanistic complementarity rather than strict uniform dominance.
This aligns with their targeted failure modes (see Section~\ref{sec:results_dynamics}): stability diagnostics show DA mitigates homogeneous-group collapse, PT maintains competitive rewards with controlled KL drift, and MT increases within-group diversity.
Consequently, rather than a single ablation dominating everywhere, the best-performing variant depends on the model and benchmark.
For example, on Gemma-2-2B, PT-GRPO yields the strongest strict success (HSR $65.8\%$ vs.\ $62.4\%$ for GRPO). On Llama-3.2-3B, DA-PT-GRPO achieves the peak soft success rate (SSR $71.5\%$), slightly edging out the full MDP-GRPO pipeline ($71.3\%$).
The full pipeline yields the most consistent stability-performance profile across all benchmarks, even if it does not always achieve the peak score on any single metric.

On the Custom Test Set, which emphasizes difficult constraint compositions, anchoring proves most consistently beneficial.
For example, DA-GRPO improves Gemma-2-2B strict success from $29.0\%$ (GRPO) to $32.6\%$, consistent with mitigating zero-variance collapse.

\textbf{Impact of Group Size and Compute Trade-offs.}
To further analyze the role of multi-temperature sampling, we evaluated both models using a restricted group size of $G=4$ (Table~\ref{tab:g4_results}).
In this low-diversity regime, the gains from MT are significantly magnified.
For Gemma-2-2B, MT-GRPO improves over standard GRPO by $+1.4\% / +1.3\%$ (SSR/HSR) on IFEval, and $+1.1\% / \mathbf{+2.0\%}$ on the Custom Test Set. Notably, MT-GRPO at $G=4$ approaches the performance of GRPO at $G=8$ (e.g., attaining an HSR of $30.6\%$ vs.\ $29.0\%$ for $G=8$ GRPO on the Custom set). Furthermore, when averaging across benchmarks under $G=4$, the full MDP-GRPO pipeline achieves the highest overall performance, whereas DA-PT-GRPO yielded the best average under $G=8$. This regime dependence confirms that MT is crucial for recovering performance under tight compute budgets by synthetically injecting within-group diversity.

\begin{figure}[t]
\includegraphics[trim={0cm 0.4cm 0cm 0.4cm}, width=0.96\linewidth]{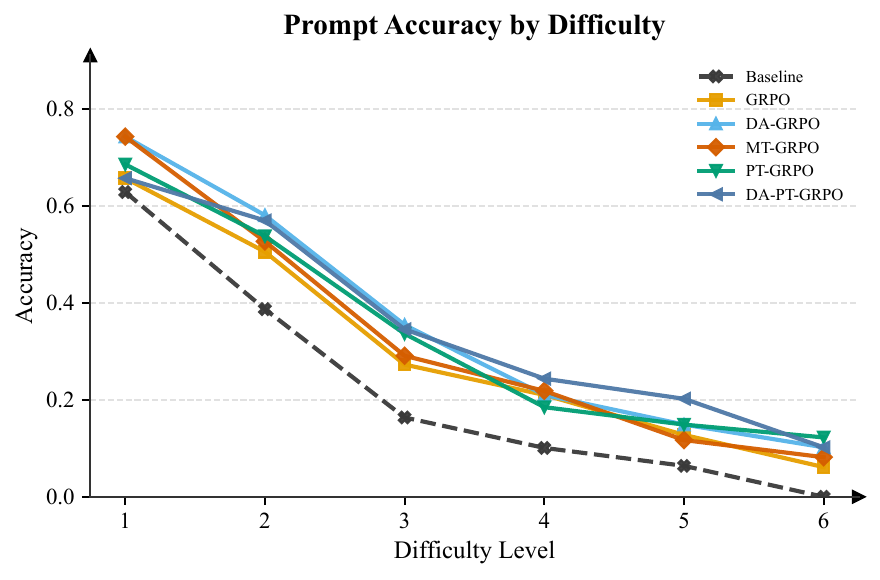}
\centering
\caption{Prompt accuracy (HSR) by difficulty, defined by the number of constraints per prompt (1--6).}
\label{fig:difficulty_promptacc}
\end{figure}

\begin{figure}[t]
\includegraphics[trim={-0.6cm 0.4cm 0cm 0.4cm}, width=0.92\linewidth]{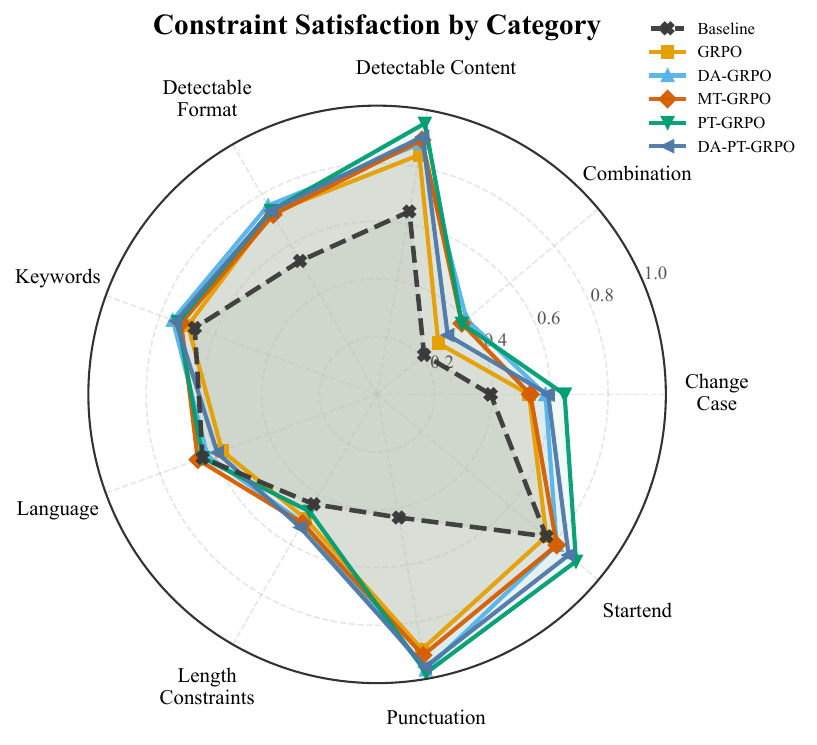}
\centering
\caption{Constraint satisfaction by category. Each spoke reports instruction-level satisfaction (SSR).}
\label{fig:category_breakdown}
\end{figure}

\begin{figure*}[t]
  \centering
  \begin{subfigure}[t]{0.49\textwidth}
    \centering
    \includegraphics[width=\linewidth]{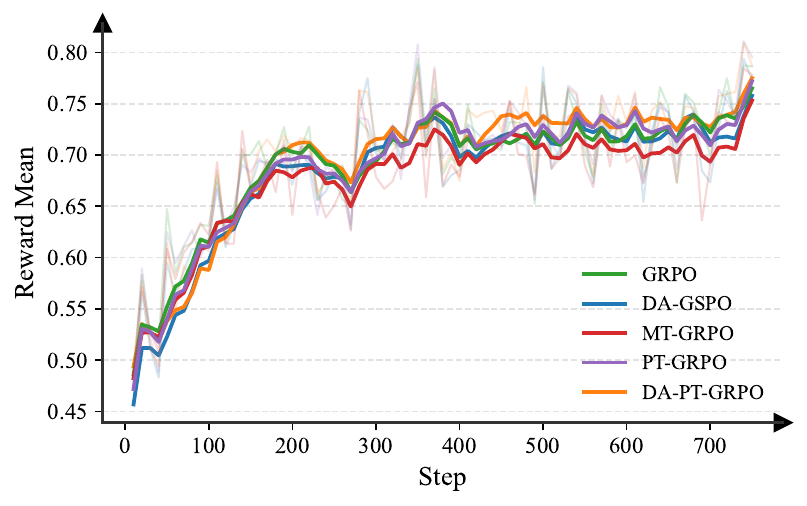}
    \vspace{-0.6cm}
    \caption{Training reward mean}
    \label{fig:diag_reward_mean}
  \end{subfigure}\hfill
  \begin{subfigure}[t]{0.49\textwidth}
    \centering
    \includegraphics[width=\linewidth]{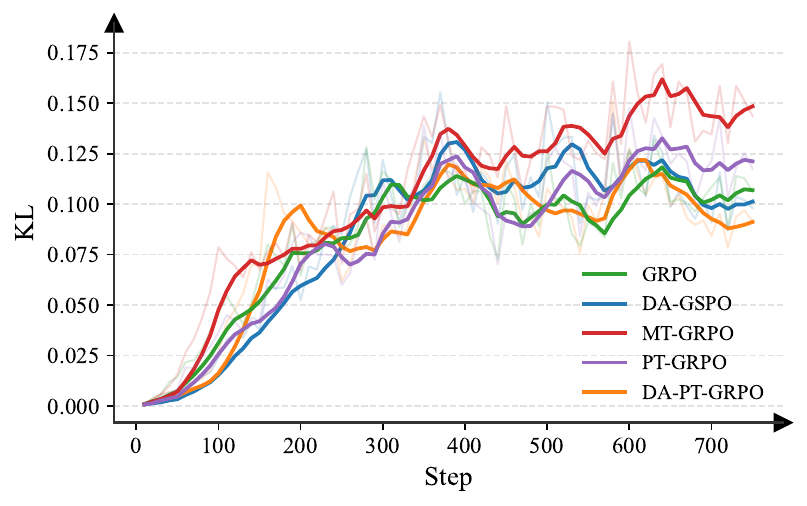}
    \vspace{-0.6cm}
    \caption{KL divergence during training}
    \label{fig:diag_kl}
  \end{subfigure}
  \begin{subfigure}[t]{0.49\textwidth}
    \centering
    \includegraphics[width=\linewidth]{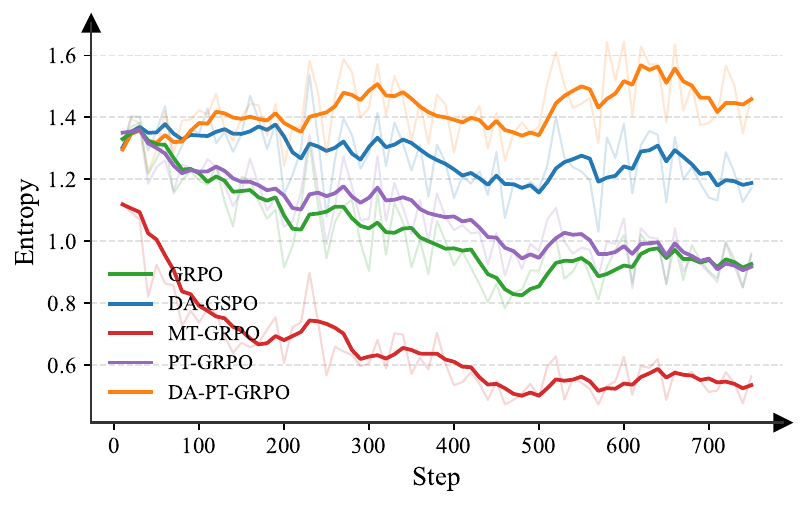}
    \vspace{-0.6cm}
    \caption{Policy entropy over training}
    \label{fig:diag_entropy}
  \end{subfigure}\hfill
  \vspace{-0.2cm}
  \begin{subfigure}[t]{0.49\textwidth}
    \centering
    \includegraphics[width=\linewidth]{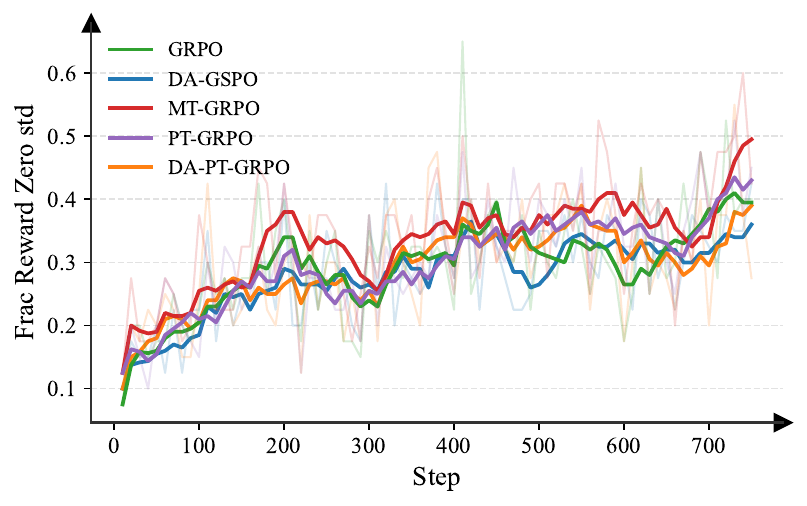}
    \vspace{-0.6cm}
    \caption{Homogeneous-group rate}
    \label{fig:diag_frac_zero_std}
  \end{subfigure}
  \caption{Training dynamics and stability diagnostics (G=8).
  (a) Mean verifiable reward,
  (b) KL divergence to the reference policy,
  (c) policy entropy, and (d) fraction of groups with zero within-group reward standard deviation.}
  \label{fig:training_diagnostics}
\end{figure*}

\subsection{Performance by Difficulty and Category}
\label{subsec:fine_grained}

Figure~\ref{fig:difficulty_promptacc} stratifies strict success by the number of constraints.
All methods degrade with difficulty, but stabilized variants degrade more slowly in the high-complexity regime: the unaligned baseline drops below 10\% HSR by Difficulty 4, while DA-PT-GRPO retains roughly 20\% HSR at Difficulty 5 (vs.\ $\approx$12\% for GRPO). This matches the intended role of anchoring and shaping in preventing near-zero learning signals when feasible solutions are rare.

Figure~\ref{fig:category_breakdown} shows that improvements are concentrated on deterministic, verifiable compliance constraints (e.g., format/content, punctuation, length and boundary constraints), where RLVR drives satisfaction toward near-ceiling levels. In contrast, gains are limited for constraints that depend primarily on semantic knowledge (e.g., language), suggesting that verifiable rewards are most effective at correcting compliance failures rather than adding new knowledge.

\subsection{Training Dynamics and Failure-Mode Diagnostics}
\label{sec:results_dynamics}

Figure~\ref{fig:training_diagnostics} connects training behavior to our targeted failure modes.
DA-GRPO reduces \texttt{frac\_reward\_zero\_std}, indicating fewer effectively homogeneous groups and supporting its role in mitigating collapse under discrete multi-constraint rewards.
PT-GRPO achieves comparable reward while maintaining a controlled KL divergence trajectory (and DA-PT-GRPO further suppresses KL drift), consistent with bounded, loss-averse shaping stabilizing updates.
MT-GRPO can increase KL and reduce entropy under our current schedule, motivating careful decoding control and temperature tuning.

\subsection{Summary of Findings}
\label{sec:results_summary}

Our findings indicate that (i) standard GRPO substantially improves verifiable instruction following over the zero-shot baseline; (ii) dual anchoring mitigates homogeneous-group collapse, yielding higher strict success on complex prompts; and (iii) prospect-theoretic shaping improves training stability by controlling KL drift while preserving reward gains, with multi-temperature sampling providing additional benefits in low-diversity regimes.

\section{Conclusion}
We investigated reinforcement learning with verifiable rewards for multi-constraint instruction following, where rewards are discrete and sometimes low-variance.
We identify three failure modes of standard group-relative advantages and propose MDP-GRPO, which combines temperature-diverse group sampling, dual-anchor advantages, and prospect-theoretic shaping.
Across IFEval, FollowBench, and a custom multi-constraint evaluation set, MDP-GRPO variants consistently improved both soft and hard success rates, achieving the most significant gains on more complex instructions.
Training diagnostics confirm that our modifications successfully reduce homogeneous-group collapse and control KL drift, ultimately enhancing alignment without severely degrading general capabilities.

\section*{Limitations}
Our methodology relies on explicit constraints verified by deterministic automatic checkers. While this ensures reliable and computationally inexpensive rewards, it does not naturally extend to subjective, stylistic, or underspecified constraints. Applying MDP-GRPO to such open-ended scenarios would likely necessitate learned reward models or preference feedback, thereby reintroducing potential reward misspecification and judge bias.

Furthermore, MDP-GRPO introduces several hyperparameters, including anchor mixing weights, shaping parameters, and temperature schedules. Although we provide sensitivity analyses for the core components, transferring this approach to vastly different domains or reward scales may require recalibration and careful monitoring of KL divergence.

Finally, our empirical validation focuses on standard instruction-following benchmarks, a custom complex constraint set, and two specific instruction-tuned model sizes (2B and 3B). Broader validation across larger model families, multilingual contexts, and highly structured domains (e.g., complex tool use or code generation) remains an area for future work to fully characterize the generalizability of our approach.

\section*{Ethical Considerations}
The primary objective in this work is improving reliability on \emph{explicit, automatically checkable} constraints (e.g., formatting and structural requirements). Our method does not, by itself, ensure broader safety properties such as harmlessness or factuality beyond what is encoded in the reward specification.

\section*{Acknowledgments}
We thank MCILab for supporting this research and providing the computational resources used in this study.

% Bibliography entries for the entire Anthology, followed by custom entries
%\bibliography{anthology,custom}
% Custom bibliography entries only
\bibliography{custom}

@article{zhou2023ifeval,
  title={{I}nstruction-{F}ollowing Evaluation for Large Language Models},
  author={Zhou, Jeffrey and Lu, Tianjian and Mishra, Swaroop and Brahma, Siddhartha and Basu, Sujoy and Luan, Yi and Zhou, Denny and Hou, Le},
  journal={arXiv preprint arXiv:2311.07911},
  year={2023},
}

@inproceedings{jiang2024followbench,
  title={{F}ollow{B}ench: A Multi-level Fine-grained Constraints Following Benchmark for Large Language Models},
  author={Jiang, Yuxin  and Wang, Yufei  and Zeng, Xingshan  and Zhong, Wanjun  and Li, Liangyou  and Mi, Fei  and Shang, Lifeng  and Jiang, Xin  and Liu, Qun  and Wang, Wei},
  booktitle = "Proceedings of the 62nd Annual Meeting of the Association for Computational Linguistics (Volume 1: Long Papers)",
  publisher = "Association for Computational Linguistics",
  year={2024}
}

@inproceedings{liu2025recast,
  title={{RECAST}: Expanding the Boundaries of {LLM}s' Complex Instruction Following with Multi-Constraint Data},
  author= {Liu, Wenhao and Guo, Zhengkang and Xie, Mingchen and Xu, Jingwen and Huang, Zisu and Tian, Muzhao and Xu, Jianhan and Wu, Muling and Wang, Xiaohua and Lv, Changze and Wang, He-Da and Yao, Hu and Zheng, Xiaoqing and Huang, Xuanjing},
  booktitle={Submitted to The Fourteenth International Conference on Learning Representations},
  year={2025},
  note={under review}
}

@article{ouyang2022instructgpt,
  title={Training language models to follow instructions with human feedback},
  author={Ouyang, Long and Wu, Jeff and Jiang, Xu and Almeida, Diogo and Wainwright, Carroll and Mishkin, Pamela and others},
  journal={Advances in neural information processing systems},
  year={2022}
}

@article{shao2024deepseekmath,
  title={{DeepSeekMath}: Pushing the Limits of Mathematical Reasoning in Open Language Models},
  author={Zhihong Shao and Peiyi Wang and Qihao Zhu and Runxin Xu and Junxiao Song and Xiao Bi and Haowei Zhang and Mingchuan Zhang and Y. K. Li and Y. Wu and Daya Guo},
  year={2024},
  journal={arXiv preprint arXiv:2402.03300},
}

@article{guo2025deepseekr1,
  title={{DeepSeek-R1} incentivizes reasoning in LLMs through reinforcement learning},
  author={Guo, Daya and Yang, Dejian and Zhang, Haowei and Song, Junxiao and Wang, Peiyi and Zhu, Qihao and Xu, Runxin and Zhang, Ruoyu and Ma, Shirong and Bi, Xiao and Zhang, Xiaokang and Yu, Xingkai and Wu, Yu and Wu, Z. F. and Gou, Zhibin and Shao, Zhihong and Li, Zhuoshu and Gao, Ziyi and Liu, Aixin and Xue, Bing and Wang, Bingxuan and Wu, Bochao and Feng, Bei and Lu, Chengda and Zhao, Chenggang and Deng, Chengqi and Ruan, Chong and Dai, Damai and Chen, Deli and Ji, Dongjie and Li, Erhang and Lin, Fangyun and Dai, Fucong and Luo, Fuli and Hao, Guangbo and Chen, Guanting and Li, Guowei and Zhang, H. and Xu, Hanwei and Ding, Honghui and Gao, Huazuo and Qu, Hui and Li, Hui and Guo, Jianzhong and Li, Jiashi and Chen, Jingchang and Yuan, Jingyang and Tu, Jinhao and Qiu, Junjie and Li, Junlong and Cai, J. L. and Ni, Jiaqi and Liang, Jian and Chen, Jin and Dong, Kai and Hu, Kai and You, Kaichao and Gao, Kaige and Guan, Kang and Huang, Kexin and Yu, Kuai and Wang, Lean and Zhang, Lecong and Zhao, Liang and Wang, Litong and Zhang, Liyue and Xu, Lei and Xia, Leyi and Zhang, Mingchuan and Zhang, Minghua and Tang, Minghui and Zhou, Mingxu and Li, Meng and Wang, Miaojun and Li, Mingming and Tian, Ning and Huang, Panpan and Zhang, Peng and Wang, Qiancheng and Chen, Qinyu and Du, Qiushi and Ge, Ruiqi and Zhang, Ruisong and Pan, Ruizhe and Wang, Runji and Chen, R. J. and Jin, R. L. and Chen, Ruyi and Lu, Shanghao and Zhou, Shangyan and Chen, Shanhuang and Ye, Shengfeng and Wang, Shiyu and Yu, Shuiping and Zhou, Shunfeng and Pan, Shuting and Li, S. S. and Zhou, Shuang and Wu, Shaoqing and Yun, Tao and Pei, Tian and Sun, Tianyu and Wang, T. and Zeng, Wangding and Liu, Wen and Liang, Wenfeng and Gao, Wenjun and Yu, Wenqin and Zhang, Wentao and Xiao, W. L. and An, Wei and Liu, Xiaodong and Wang, Xiaohan and Chen, Xiaokang and Nie, Xiaotao and Cheng, Xin and Liu, Xin and Xie, Xin and Liu, Xingchao and Yang, Xinyu and Li, Xinyuan and Su, Xuecheng and Lin, Xuheng and Li, X. Q. and Jin, Xiangyue and Shen, Xiaojin and Chen, Xiaosha and Sun, Xiaowen and Wang, Xiaoxiang and Song, Xinnan and Zhou, Xinyi and Wang, Xianzu and Shan, Xinxia and Li, Y. K. and Wang, Y. Q. and Wei, Y. X. and Zhang, Yang and Xu, Yanhong and Li, Yao and Zhao, Yao and Sun, Yaofeng and Wang, Yaohui and Yu, Yi and Zhang, Yichao and Shi, Yifan and Xiong, Yiliang and He, Ying and Piao, Yishi and Wang, Yisong and Tan, Yixuan and Ma, Yiyang and Liu, Yiyuan and Guo, Yongqiang and Ou, Yuan and Wang, Yuduan and Gong, Yue and Zou, Yuheng and He, Yujia and Xiong, Yunfan and Luo, Yuxiang and You, Yuxiang and Liu, Yuxuan and Zhou, Yuyang and Zhu, Y. X. and Huang, Yanping and Li, Yaohui and Zheng, Yi and Zhu, Yuchen and Ma, Yunxian and Tang, Ying and Zha, Yukun and Yan, Yuting and Ren, Z. Z. and Ren, Zehui and Sha, Zhangli and Fu, Zhe and Xu, Zhean and Xie, Zhenda and Zhang, Zhengyan and Hao, Zhewen and Ma, Zhicheng and Yan, Zhigang and Wu, Zhiyu and Gu, Zihui and Zhu, Zijia and Liu, Zijun and Li, Zilin and Xie, Ziwei and Song, Ziyang and Pan, Zizheng and Huang, Zhen and Xu, Zhipeng and Zhang, Zhongyu and Zhang, Zhen},
  journal={Nature},
  year={2025}
}

@article{wen2025rlvr,
  title={{Reinforcement Learning with Verifiable Rewards Implicitly Incentivizes Correct Reasoning in Base LLMs}},
  author={Wen, Xumeng and Liu, Zihan and Zheng, Shun and Xu, Zhijian and Ye, Shengyu and others},
  year={2025},
  journal={arXiv preprint arXiv:2506.14245},
}

@inproceedings{ethayarajh2024kto,
  title     = {{KTO: Model Alignment as Prospect Theoretic Optimization}},
  author    = {Ethayarajh, Kawin and Xu, Winnie and Muennighoff, Niklas and Jurafsky, Dan and Kiela, Douwe},
  booktitle = {Proceedings of the 41st International Conference on Machine Learning (ICML)},
  year      = {2024},
  publisher = {PMLR},
}

@article{chung2022instructiontuning,
  title={Scaling Instruction-Finetuned language models},
  author={Chung, Hyung Won and Hou, Le and Longpre, Shayne and Zoph, Barret and others},
  journal={Journal of Machine Learning Research},
  year={2024}
}

@inproceedings{chen2024judgebias,
  title={{Humans or LLMs as the Judge? A Study on Judgement Bias}},
  author={Guiming Hardy Chen and Shunian Chen and Ziche Liu and Feng Jiang and Benyou Wang},
  booktitle = "Proceedings of the 2024 Conference on Empirical Methods in Natural Language Processing",
  publisher = "Association for Computational Linguistics",
  year={2024}
}

@inproceedings{ye2025justiceprejudice,
  title={{Justice or Prejudice? Quantifying Biases in LLM-as-a-Judge}},
  author={Jiayi Ye and Yanbo Wang and Yue Huang and Dongping Chen and Qihui Zhang and Nuno Moniz and Tian Gao and Werner Geyer and Chao Huang and Pin-Yu Chen and Nitesh V Chawla and Xiangliang Zhang},
  booktitle={The Thirteenth International Conference on Learning Representations},
  year={2025},
}

@inproceedings{lin2023alignmenttax,
  title={{Mitigating the Alignment Tax of RLHF}},
  author = "Lin, Yong  and
      Lin, Hangyu  and
      Xiong, Wei  and
      Diao, Shizhe  and
      Liu, Jianmeng  and
      Zhang, Jipeng  and
      Pan, Rui  and
      Wang, Haoxiang  and
      Hu, Wenbin  and
      Zhang, Hanning  and
      Dong, Hanze  and
      Pi, Renjie  and
      Zhao, Han  and
      Jiang, Nan  and
      Ji, Heng  and
      Yao, Yuan  and
      Zhang, Tong",
  booktitle = "Proceedings of the 2024 Conference on Empirical Methods in Natural Language Processing",
  publisher = "Association for Computational Linguistics",
  year={2024},
}

@inproceedings{kotha2024forgetting,
  title={{Understanding Catastrophic Forgetting in Language Model Fine-tuning}},
  author={Kotha, Suhas and Springer, Jacob Mitchell and Raghunathan, Aditi},
  booktitle={NeurIPS 2023 Workshop on Distribution Shifts: New Frontiers with Foundation Models},
  year={2024},
}

@article{kahneman1979prospect,
  title={{Prospect Theory: An Analysis of Decision under Risk}},
  author={Kahneman, Daniel and Tversky, Amos},
  journal = {Econometrica},
  publisher = {[Wiley, Econometric Society]},
  year = {1979}
}

@article{hendrycks2021mmlu,
  title={{Measuring Massive Multitask Language Understanding}},
  author={Hendrycks, Dan and Burns, Collin and Basart, Steven and Zou, Andy and Mazeika, Mantas and Song, Dawn and Steinhardt, Jacob},
  journal={arXiv preprint arXiv:2009.03300},
  year={2020},
}

@article{clark2018arc,
  title={{Think you have Solved Question Answering? Try ARC, the AI2 Reasoning Challenge}},
  author={Clark, Peter and Cowhey, Isaac and Etzioni, Oren and Khot, Tushar and Sabharwal, Ashish and Schoenick, Carissa and Tafjord, Oyvind},
  journal={arXiv preprint arXiv:1803.05457},
  year={2018},
}

@article{schick2023toolformer,
  title={{Toolformer: Language Models Can Teach Themselves to Use Tools}},
  author={Schick, Timo and Dwivedi-Yu, Jane and Dess{\`\i}, Roberto and Raileanu, Roberta and Lomeli, Maria and Zettlemoyer, Luke and Cancedda, Nicola and Scialom, Thomas},
  journal={Advances in Neural Information Processing Systems},
  year={2023},
}

@inproceedings{yao2023react,
  title={{ReAct: Synergizing Reasoning and Acting in Language Models}},
  author={Yao, Shunyu and Zhao, Jeffrey and Yu, Dian and Du, Nan and Shafran, Izhak and Narasimhan, Karthik R and Cao, Yuan},
  booktitle={The eleventh international conference on learning representations},
  year={2022}
}

@inproceedings{shen2025slot,
  title={{SLOT: Structuring the Output of Large Language Models}},
  author="Shen, Zhengyuan  and
      Wang, Darren Yow-Bang  and
      Mishra, Soumya Smruti  and
      Xu, Zhichao  and
      Teng, Yifei  and
      Ding, Haibo",
  booktitle = "Proceedings of the 2025 Conference on Empirical Methods in Natural Language Processing: Industry Track",
  year = "2025",
  publisher = "Association for Computational Linguistics",
}

@article{chen2025safetyconstraints,
  title={{Learning Safety Constraints for Large Language Models}},
  author={Chen, Xin and As, Yarden and Krause, Andreas},
  journal={arXiv preprint arXiv:2505.24445},
  year={2025},
}

@inproceedings{westermann2024dallma,
  title={{Dallma: Semi-Structured Legal Reasoning and Drafting with Document Automation and LLM Assistance}},
  author={Hannes Westermann},
  booktitle={ICML Workshop (GenLaw)},
  year={2024}
}

@inproceedings{narendra2024contracts,
  title={{Enhancing Contract Negotiations with LLM-Based Legal Document Comparison}},
  author={Narendra, Savinay and Shetty, Kaushal and Ratnaparkhi, Adwait},
  booktitle={Natural Legal Language Processing Workshop},
  publisher = "Association for Computational Linguistics",
  year={2024}
}

@inproceedings{geng2023gcd,
  title={{Grammar-Constrained Decoding for Structured NLP Tasks without Finetuning}},
  author={Saibo Geng and Martin Josifoski and Maxime Peyrard and Robert West},
  booktitle = "Proceedings of the 2023 Conference on Empirical Methods in Natural Language Processing",
  publisher = "Association for Computational Linguistics",
  year={2023}
}

@article{park2025gcd,
  title={{Flexible and Efficient Grammar-Constrained Decoding}},
  author={Kanghee Park and Timothy Zhou and Loris D'Antoni},
  journal={arXiv preprint arXiv:2502.05111},
  year={2025},
}

@article{shapingsparse2025,
  title={{Shaping Sparse Rewards in Reinforcement Learning}},
  author={Li, Wenyun and Huang, Wenjie and Sun, Chen},
  journal={arXiv preprint arXiv:2501.19128},
  year={2025},
}

@article{mapo2025,
  title={{MAPO}: {M}ixed {A}dvantage {P}olicy {O}ptimization},
  author={Wenke Huang and Quan Zhang and Yiyang Fang and Jian Liang and Xuankun Rong and Huanjin Yao and Guancheng Wan and Ke Liang and Wenwen He and Mingjun Li and Leszek Rutkowski and Mang Ye and Bo Du and Dacheng Tao},
  journal={arXiv preprint arXiv:2509.18849},
  year={2025}
}

@inproceedings{he-etal-2024-complex,
    title = {{From Complex to Simple: Enhancing Multi-Constraint Complex Instruction Following Ability of Large Language Models}},
    author = {He, Qianyu and Zeng, Jie and He, Qianxi and Liang, Jiaqing and Xiao, Yanghua},
    booktitle = {Findings of the Association for Computational Linguistics: EMNLP 2024},
    year = "2024",
    address = "Miami, Florida, USA",
    publisher = "Association for Computational Linguistics",
}

@article{nan2025ngrpo,
  title={{NGRPO}: Negative-enhanced group relative policy optimization},
  author={Nan, Gongrui and Chen, Siye and Huang, Jing and Lu, Mengyu and Wang, Dexun and Xie, Chunmei and Xiong, Weiqi and Zeng, Xianzhou and Zhou, Qixuan and Li, Yadong and others},
  journal={arXiv preprint arXiv:2509.18851},
  year={2025}
}

@inproceedings{sutton1999policy,
 author = {Sutton, Richard S and McAllester, David and Singh, Satinder and Mansour, Yishay},
 booktitle = {Advances in Neural Information Processing Systems},
 editor = {S. Solla and T. Leen and K. M\"{u}ller},
 publisher = {MIT Press},
 title = {{Policy Gradient Methods for Reinforcement Learning with Function Approximation}},
 volume = {12},
 year = {1999}
}

\clearpage
\appendix

\section{MDP-GRPO Algorithm}
\label{sec:appendix_alg}

Algorithm \ref{alg:pg} details the full training loop for MDP-GRPO.
The procedure differs from standard GRPO through four key additions:
(1) sampling is performed with a heterogeneous temperature schedule $\mathbf{T}$ to ensure group diversity;
(2) absolute goal-aware anchors ($\delta_i$) are computed alongside standard group-relative scores ($z_i$);
(3) prospect-theoretic advantage shaping is applied via a bounded, asymmetric function to enforce loss aversion;
and (4) the advantages are mixed via a convex combination before being used in the PPO-style update with an asymmetric KL penalty.

\begin{algorithm}[h]
\caption{MDP-GRPO for Multi-Constraint Instruction Following}
\label{alg:pg}
\begin{algorithmic}[1]
\Require Policy $\pi_\theta$, Reference Policy $\pi_{\text{ref}}$, Temperature Schedule $\mathbf{T}$, Mixing Weight $\alpha$, Shaping Params $(\lambda_{+},\lambda_{-},\beta_{\text{PT}})$.
\For{each training step}
    \State Sample a batch of prompts $x \sim \mathcal{D}$.
    \For{each prompt $x$}
        \State Sample $G$ completions $y_i \sim \pi_{\theta_{\text{old}}}(\cdot\mid x; \tau_i)$ using schedule $\mathbf{T}$.
        \State Compute rewards $r_i$ for all $i \in \{1, \dots, G\}$. % (Eq.~\ref{eq:reward}) if you have a label
        \State Compute standard group-relative advantages $z_i$ (Eq.~\ref{eq:standard_grpo_adv}).
        \State Compute absolute goal-aware anchors $\delta_i$ (Eq.~\ref{eq:delta}).
        \State Apply Prospect-Theoretic shaping to advantages (Eq.~\ref{eq:prospect}).
        \State Compute mixed advantages (Eq.~\ref{eq:mix_advantages}).
    \EndFor
    \State Compute the surrogate loss $\mathcal{J}(\theta)$ (Eq.~\ref{eq:objective}) including the Asymmetric KL penalty.
    \State Update $\theta$ via gradient descent.
\EndFor
\end{algorithmic}
\end{algorithm}

\section{Dual-Anchoring Ablations}
\label{sec:appendix_da_ablations}

\subsection{\texorpdfstring{Effect of the dual-anchor mixing weight $\alpha$}{Effect of the dual-anchor mixing weight alpha}}

Introducing $\delta_i$ modifies the effective advantage and optimizes a surrogate objective.
This introduces a controlled bias relative to standard GRPO, analogous to other widely used modifications such as PPO clipping or advantage normalization that trade strict unbiasedness for drastically reduced variance under finite-sample noise.
This bias-variance tradeoff is continuously controlled by $\alpha$: when $\alpha = 0$, we recover standard GRPO exactly. In low-dispersion discrete-reward regimes where group normalization becomes ill-conditioned, the variance reduction from $\delta_i$ dominates.

\begin{figure*}[h]
  \centering
  \begin{subfigure}[h]{0.33\textwidth}
    \centering
    \includegraphics[trim={0.2cm 0.2cm 0.2cm 0.2cm},clip,width=\linewidth]{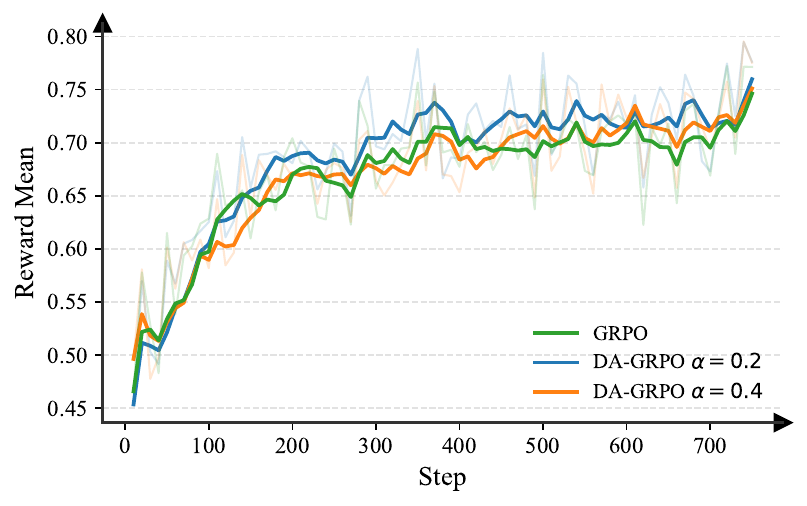}
    \caption{Training reward mean}
    \label{fig:reward_mean_smooth_DA_alpha}
  \end{subfigure}\hfill
  \begin{subfigure}[h]{0.33\textwidth}
    \centering
    \includegraphics[trim={0.2cm 0.2cm 0.2cm 0.2cm},clip,width=\linewidth]{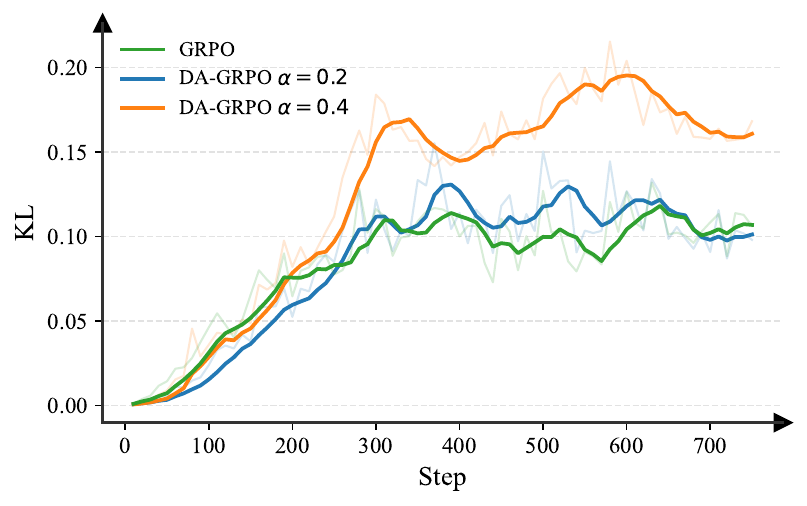}
    \caption{KL divergence during training}
    \label{fig:kl_smoother_DA_alpha}
  \end{subfigure}\hfill
  \begin{subfigure}[h]{0.33\textwidth}
    \centering
    \includegraphics[trim={0.2cm 0.2cm 0.2cm 0.2cm},clip,width=\linewidth]{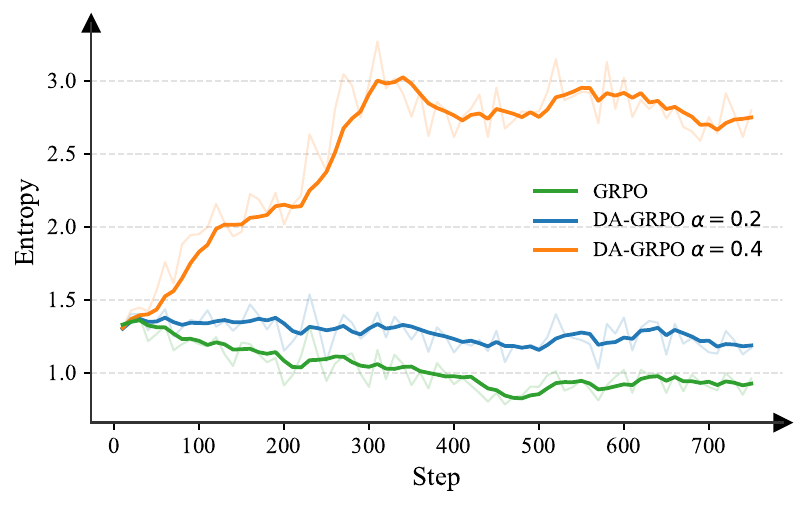}
    \caption{Policy entropy over training}
    \label{fig:entropy_smoother_DA_alpha}
  \end{subfigure}
  \caption{Dual-anchor mixing weight $\alpha$ sweep.
Training dynamics for GRPO and dual-anchor GRPO (DA-GRPO) with $\alpha\!\in\!\{0.2,0.4\}$.
Panels show (a) mean verifiable reward, (b) KL divergence to the reference policy, and (c) policy entropy, highlighting that larger $\alpha$ induces more aggressive policy updates (higher KL/entropy) with limited additional reward gain.}

\end{figure*}

\Cref{fig:entropy_smoother_DA_alpha} compares GRPO with dual-anchoring (DA-GRPO) for $\alpha\in\{0.2,0.4\}$.
Across training, the mean verifiable reward rises similarly for all settings and converges to comparable final values, indicating that dual-anchoring does not hinder optimization.
However, increasing $\alpha$ noticeably increases the KL divergence: $\alpha=0.4$ produces substantially larger KL throughout training, suggesting more aggressive policy shifts driven by the absolute (anchor-based) component.
Since $\alpha=0.4$ does not yield commensurate reward gains but incurs higher KL (and thus stronger deviation from the reference policy), we use a moderate value ($\alpha=0.2$) as a better reward--stability trade-off.

\subsection{\texorpdfstring{Sensitivity to the goal-aware center used in $\delta$}{Sensitivity to the goal-aware center used in delta}}

\begin{figure*}[h]
  \centering
  \begin{subfigure}[h]{0.33\textwidth}
    \centering
    \includegraphics[trim={0.2cm 0.2cm 0.2cm 0.2cm},clip,width=\linewidth]{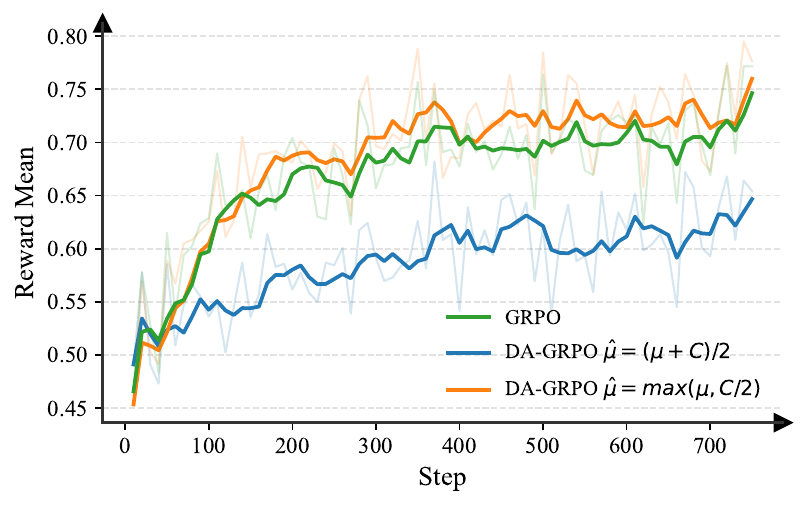}
    \caption{Training reward mean}
    \label{fig:reward_mean_smooth_DA_mean}
  \end{subfigure}\hfill
  \begin{subfigure}[h]{0.33\textwidth}
    \centering
    \includegraphics[trim={0.2cm 0.2cm 0.2cm 0.2cm},clip,width=\linewidth]{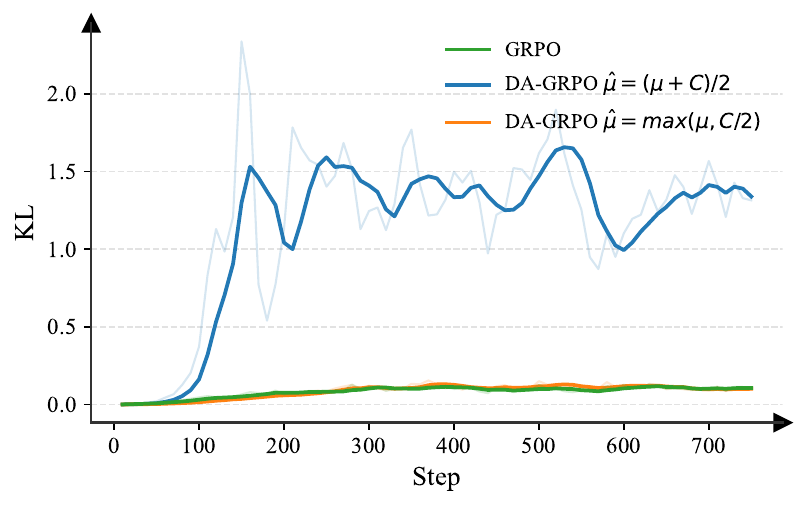}
    \caption{KL divergence during training}
    \label{fig:kl_smoother_DA_mean}
  \end{subfigure}
  \begin{subfigure}[h]{0.33\textwidth}
    \centering
    \includegraphics[trim={0.2cm 0.2cm 0.2cm 0.2cm},clip,width=\linewidth]{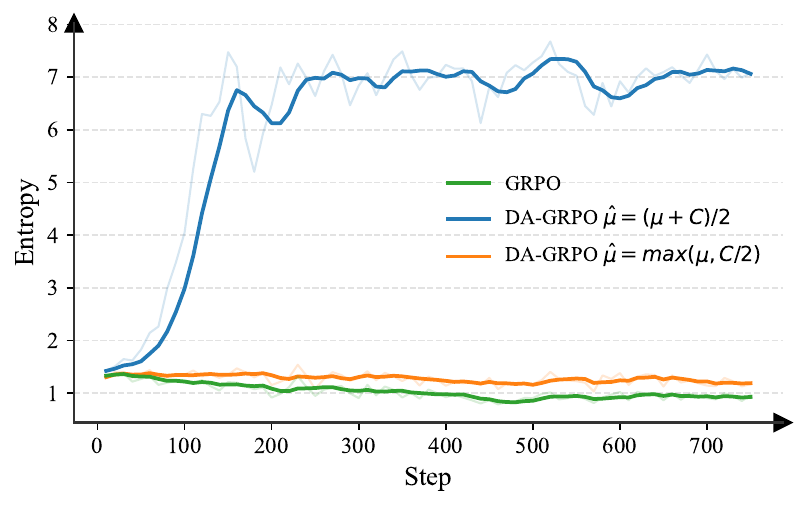}
    \caption{Policy entropy over training}
    \label{fig:entropy_smoother_DA_mean}
  \end{subfigure}
  \caption{Sensitivity to the goal-aware center in the anchor term.
Comparison of DA-GRPO under two choices of the anchor center used in $\delta_i$:
$\hat{\mu}=(\mu_{\text{group}}+C)/2$ versus $\hat{\mu}=\max(\mu_{\text{group}},\,C/2)$ (for normalized rewards, $\max(\mu_{\text{group}},0.5)$).
Panels report (a) mean verifiable reward, (b) KL divergence to the reference policy, and (c) policy entropy, showing that the conservative piecewise center preserves stability while the upward-shifted center induces large drift and elevated entropy without improving reward.}

\end{figure*}

\Cref{fig:entropy_smoother_DA_mean} compares two choices for the anchor center in the goal-aware term:
(i) $\hat{\mu}=(\mu+C)/2$, which shifts the reference point upward as training progresses, and
(ii) $\hat{\mu}=\max(\mu,\,C/2)$ (for $r\in[0,1]$, this is $\max(\mu,0.5)$), which uses the neutral binomial mean unless the group is already above it.

The results show that $\hat{\mu}=(\mu+C)/2$ is unstable: it produces a large and sustained increase in KL divergence and a sharp rise in policy entropy, while achieving substantially \emph{lower} reward than both GRPO and the alternative anchor.
This suggests that aggressively moving the center upward can induce overly strong (and effectively noisy) updates that drift far from the reference policy without yielding commensurate improvement in constraint satisfaction.

In contrast, $\hat{\mu}=\max(\mu,C/2)$ tracks GRPO closely in KL and entropy (remaining near the reference policy) while matching or slightly improving the reward trajectory.
Overall, using a conservative, piecewise center that falls back to the neutral binomial mean appears to provide a better stability--performance trade-off than continuously increasing the anchor target.

\section{Prospect-theoretic Shaping and Asymmetric KL Ablations}
\label{sec:appendix_pt_ablations}

\begin{figure*}[h]
  \centering
  \begin{subfigure}[h]{0.48\textwidth}
    \centering
    \includegraphics[trim={0.2cm 0.2cm 0.2cm 0.2cm},clip,width=\linewidth]{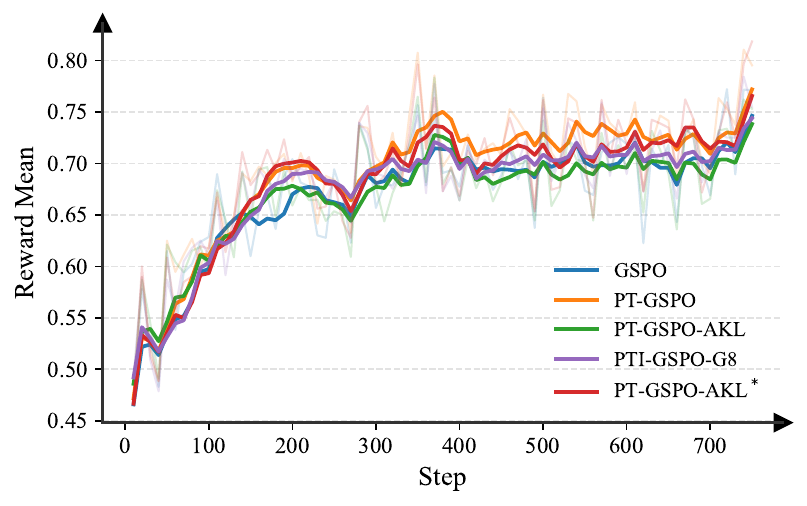}
    \caption{Training reward mean}
    \label{fig:reward_mean_smooth_PT}
  \end{subfigure}\hfill
  \begin{subfigure}[h]{0.48\textwidth}
    \centering
    \includegraphics[trim={0.2cm 0.2cm 0.2cm 0.2cm},clip,width=\linewidth]{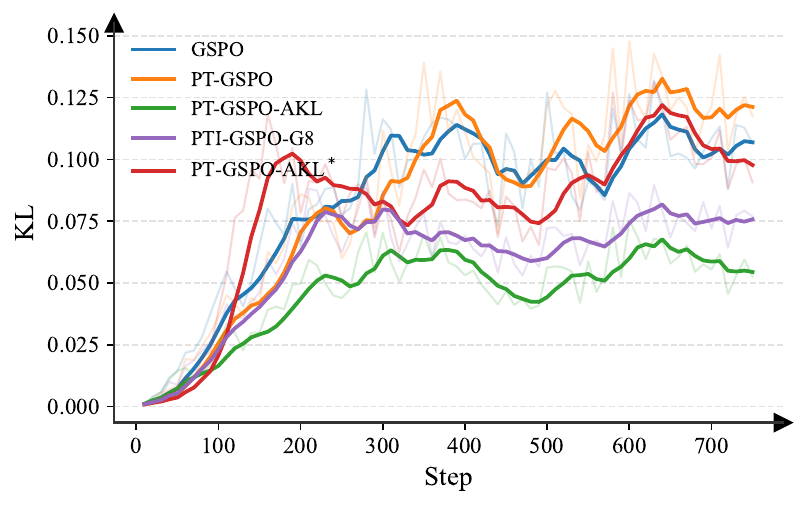}
    \caption{KL divergence during training}
    \label{fig:kl_smoother_PT}
  \end{subfigure}
  \begin{subfigure}[h]{0.48\textwidth}
    \centering
    \includegraphics[trim={0.2cm 0.2cm 0.2cm 0.2cm},clip,width=\linewidth]{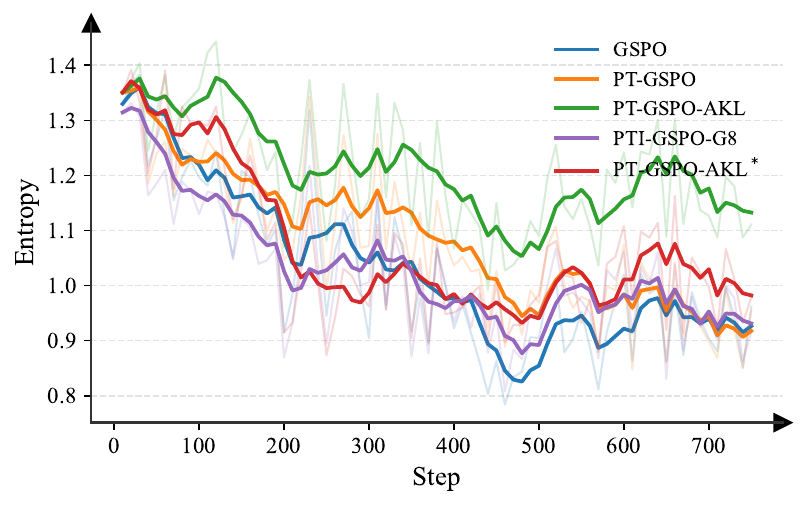}
    \caption{Policy entropy over training}
    \label{fig:entropy_smoother_PT}
  \end{subfigure}\hfill
  \begin{subfigure}[h]{0.48\textwidth}
    \centering
    \includegraphics[width=\linewidth]{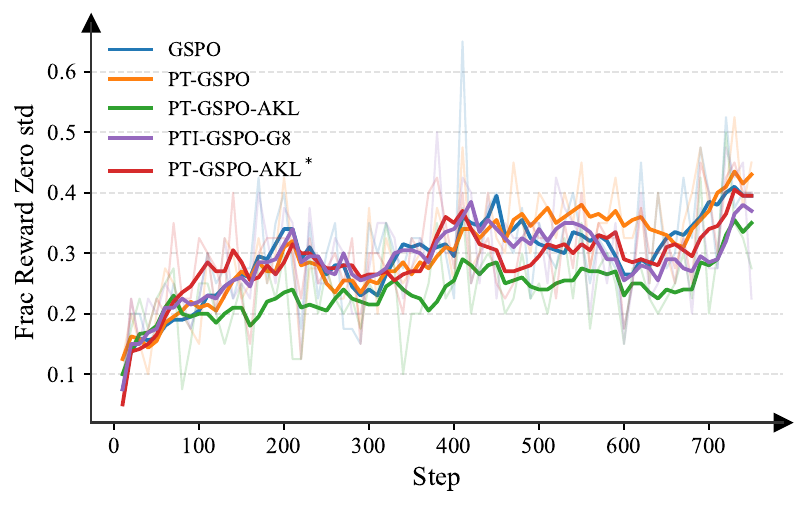}
    \caption{Frac Reward Zero std}
    \label{fig:diag_frac_zero_std_PT}
  \end{subfigure}
  \caption{Ablation of prospect-theoretic shaping and asymmetric KL under GRPO.
We compare GRPO with loss-averse shaping (PT-GRPO), its inverse variant (PTI-GRPO), and two asymmetric-KL versions (PT-GRPO-AKL / PT-GRPO-AKL$^\ast$; see text for hyperparameters).
Panels report (a) mean verifiable reward, (b) KL divergence to the reference policy, (c) policy entropy, and (d) the fraction of steps with near-zero within-group reward standard deviation (homogeneous groups), illustrating the stability--drift trade-offs induced by shaping and KL control.}

\end{figure*}

\Cref{fig:diag_frac_zero_std_PT} studies how loss-averse shaping and its variants affect training dynamics under GRPO.
Across settings, the training reward mean is broadly comparable, with PT-GRPO matching or slightly improving over the GRPO baseline, while the \emph{inverse} shaping (PTI-GRPO; $\lambda_{+}>\lambda_{-}$) does not yield consistent gains, suggesting that improvements are driven by \emph{loss aversion} rather than merely applying a bounded nonlinearity.

In terms of stability, PT-GRPO tends to increase KL relative to GRPO, indicating more aggressive policy updates when amplifying negative advantages.
Introducing asymmetric KL regularization (PT-GRPO-AKL) substantially reduces KL throughout training while maintaining competitive reward, and also preserves higher policy entropy, consistent with better-controlled updates and reduced premature collapse.
The more aggressive variant (PT-GRPO-AKL$^\ast$) partially recovers the higher-KL behavior without commensurate reward improvements, highlighting a trade-off between correction strength and policy drift.

Finally, PT-GRPO-AKL yields the lowest fraction of steps with near-zero within-group reward variance (\emph{Frac Reward Zero std}), indicating fewer effectively homogeneous groups and a more reliable learning signal over training.
Overall, loss-averse shaping is most effective when paired with stronger KL control on negative-advantage updates, yielding a better reward--stability trade-off than either shaping alone or its inverse.
\section{General Capability Retention}
\label{sec:results_general}

To test whether instruction-following gains come at the cost of general capabilities, we evaluate the models on \textsc{MMLU}, \textsc{ARC-Easy}, and \textsc{ARC-Challenge}.
\Cref{tab:mmlu_results} reports accuracy on MMLU under the zero-shot protocol, while \Cref{tab:arc_results} details performance on the ARC datasets.
Overall, our best-performing instruction-following variants preserve (or minimally change) general capability performance. This indicates that the instruction-following improvements are not achieved at the expense of general knowledge or driven by uncontrolled policy drift.

\begin{table*}[h]
\centering
\small
\setlength{\tabcolsep}{2pt}
\begin{tabular}{lccccc||ccccc}
\toprule
\multirow{2}{*}{\textbf{Models}} & \multicolumn{5}{c}{\textbf{Gemma-2-2B-Instruct}} & \multicolumn{5}{c}{\textbf{Llama-3.2-3B-Instruct}} \\
\cmidrule(lr){2-6}\cmidrule(lr){7-11}
& \multicolumn{5}{c}{\textbf{MMLU}} & \multicolumn{5}{c}{\textbf{MMLU}} \\
\cmidrule(lr){2-6}\cmidrule(lr){7-11}
& \textbf{Overall} & \textbf{Humanities} & \textbf{Social Sci.} & \textbf{STEM} & \textbf{Other} & \textbf{Overall} & \textbf{Humanities} & \textbf{Social Sci.} & \textbf{STEM} & \textbf{Other} \\
\midrule
Base & 56.9 & 50.8 & 67.3 & 48.4 & 64.3 & 60.4 & 59.2 & 67.0 & 50.3 & 65.9 \\
GRPO & 57.1 & 51.0 & 67.4 & 48.8 & 64.7 & 62.3 & 61.1 & 68.4 & 52.0 & 68.5 \\
MT-GRPO & 57.0 & 50.9 & 67.4 & 48.4 & 64.5 & 62.4 & 61.2 & 68.6 & 52.1 & 68.5 \\
DA-GRPO & 57.0 & 51.2 & 67.1 & 48.8 & 64.2 & 62.2 & 61.0 & 68.7 & 51.9 & 68.2 \\
PT-GRPO & 56.9 & 50.8 & 67.3 & 48.4 & 64.2 & 62.4 & 61.4 & 68.7 & 52.0 & 68.4 \\
DA-PT-GRPO & 56.8 & 50.8 & 67.2 & 48.2 & 64.3 & 62.1 & 60.9 & 68.7 & 51.5 & 68.1 \\
MDP-GRPO & 57.0 & 50.9 & 67.4 & 48.7 & 64.5 & 62.3 & 61.0 & 68.8 & 51.8 & 68.3 \\
\bottomrule
\end{tabular}
\caption{MMLU performance across different model configurations. Scores are reported as percentages.}
\label{tab:mmlu_results}
\end{table*}
\begin{table*}[h]
\centering
\small
\setlength{\tabcolsep}{4pt}
\begin{tabular}{lcccc||cccc}
\toprule
\multirow{2}{*}{\textbf{Models}} & \multicolumn{4}{c}{\textbf{Gemma-2-2B-Instruct}} & \multicolumn{4}{c}{\textbf{Llama-3.2-3B-Instruct}} \\
\cmidrule(lr){2-5}\cmidrule(lr){6-9}
& \multicolumn{2}{c}{\textbf{ARC-Challenge}} & \multicolumn{2}{c}{\textbf{ARC-Easy}} & \multicolumn{2}{c}{\textbf{ARC-Challenge}} & \multicolumn{2}{c}{\textbf{ARC-Easy}} \\
\cmidrule(lr){2-3}\cmidrule(lr){4-5}\cmidrule(lr){6-7}\cmidrule(lr){8-9}
& \textbf{Acc} & \textbf{Acc Norm} & \textbf{Acc} & \textbf{Acc Norm} & \textbf{Acc} & \textbf{Acc Norm} & \textbf{Acc} & \textbf{Acc Norm} \\
\midrule
Base & 50.9 & 52.5 & 81.0 & 78.3 & 43.6 & 45.6 & 74.2 & 68.8 \\
GRPO & 50.4 & 52.0 & 80.3 & 77.0 & 43.5 & 45.6 & 74.3 & 68.3 \\
MT-GRPO & 51.3 & 53.2 & 81.1 & 78.4 & 43.5 & 46.7 & 75.5 & 71.4 \\
DA-GRPO & 50.4 & 54.1 & 80.8 & 77.7 & 43.2 & 46.5 & 74.2 & 68.4 \\
PT-GRPO & 51.5 & 52.4 & 80.7 & 78.4 & 43.6 & 45.7 & 74.0 & 68.3 \\
DA-PT-GRPO & 51.0 & 53.2 & 81.6 & 78.4 & 43.0 & 46.5 & 75.5 & 71.2\\
MDP-GRPO & 50.9 & 52.5 & 81.1 & 78.2 & 43.9 & 46.5 & 75.4 & 71.8 \\
\bottomrule
\end{tabular}
\caption{ARC-Challenge and ARC-Easy performance across different model configurations. Scores are reported as percentages.}
\label{tab:arc_results}
\end{table*}

\end{document}